\documentclass{article}
\usepackage{times}
\usepackage{graphicx} % more modern
\usepackage{subfigure}
\usepackage{bm}
\usepackage{natbib}
\usepackage{algorithm}
\usepackage{algorithmic}
\usepackage{amsmath}
\usepackage{array}
\usepackage{hyperref}
\usepackage[accepted]{icml2016}
\newcommand{\qihao}{\fontsize{8pt}{\baselineskip}\selectfont}
\renewcommand\thefootnote{}

\icmltitlerunning{Large-Margin Softmax Loss for Convolutional Neural Networks}

\begin{document}

\twocolumn[
\icmltitle{Large-Margin Softmax Loss for Convolutional Neural Networks}

\icmlauthor{Weiyang Liu\textsuperscript{1\textdagger}}{wyliu@pku.edu.cn}
\icmlauthor{Yandong Wen\textsuperscript{2\textdagger}}{wen.yandong@mail.scut.edu.cn}
\icmlauthor{Zhiding Yu\textsuperscript{3}}{yzhiding@andrew.cmu.edu}
\icmlauthor{Meng Yang\textsuperscript{4}}{yang.meng@szu.edu.cn}
\icmladdress{\textsuperscript{1}School of ECE, Peking University\ \ \ \textsuperscript{2}School of EIE, South China University of Technology \\\textsuperscript{3}Dept. of ECE, Carnegie Mellon University\ \ \ \textsuperscript{4}College of CS \& SE, Shenzhen University}
\icmlkeywords{Large-Margin, Softmax, Convolutional Neural Networks, Classification}
\vskip 0.3in
]

\begin{abstract}
Cross-entropy loss together with softmax is arguably one of the most common used supervision components in convolutional neural networks (CNNs). Despite its simplicity, popularity and excellent performance, the component does not explicitly encourage discriminative learning of features. In this paper, we propose a generalized large-margin softmax (L-Softmax) loss which explicitly encourages intra-class compactness and inter-class separability between learned features. Moreover, L-Softmax not only can adjust the desired margin but also can avoid overfitting. We also show that the L-Softmax loss can be optimized by typical stochastic gradient descent. Extensive experiments on four benchmark datasets demonstrate that the deeply-learned features with L-softmax loss become more discriminative, hence significantly boosting the performance on a variety of visual classification and verification tasks.
\end{abstract}

\section{Introduction}
Over the past several years, convolutional neural networks (CNNs) have significantly boosted the state-of-the-art performance in many visual classification tasks such as object recognition, \cite{krizhevsky2012imagenet,sermanet2013overfeat,he2015delving,he2015deep}, face verification \cite{taigman2014deepface,sun2014deep,sun2015deeply} and hand-written digit recognition \cite{wan2013regularization}. The layered learning architecture, together with convolution and pooling which carefully extract features from local to global, renders the strong visual representation ability of CNNs as well as their current significant positions in large-scale visual recognition tasks. Facing the increasingly more complex data, CNNs have continuously been improved with deeper structures \cite{simonyan2014very,szegedy2015going}, smaller strides \cite{simonyan2014very} and new non-linear activations \cite{goodfellow2013maxout,nair2010rectified,he2015delving}. While benefiting from the strong learning ability, CNNs also have to face the crucial issue of overfilling. Considerable effort such as large-scale training data \cite{russakovsky2014imagenet}, dropout \cite{krizhevsky2012imagenet}, data augmentation \cite{krizhevsky2012imagenet,szegedy2015going}, regularization \cite{hinton2012improving,srivastava2014dropout,wan2013regularization,goodfellow2013maxout} and stochastic pooling \cite{zeiler2013stochastic} has been put to address the issue.\footnote{\textsuperscript{\textdagger}Authors contributed equally. Code is available at \url{https://github.com/wy1iu/LargeMargin_Softmax_Loss}}

\begin{figure}[t]
  \centering
  \includegraphics[width=2.8in]{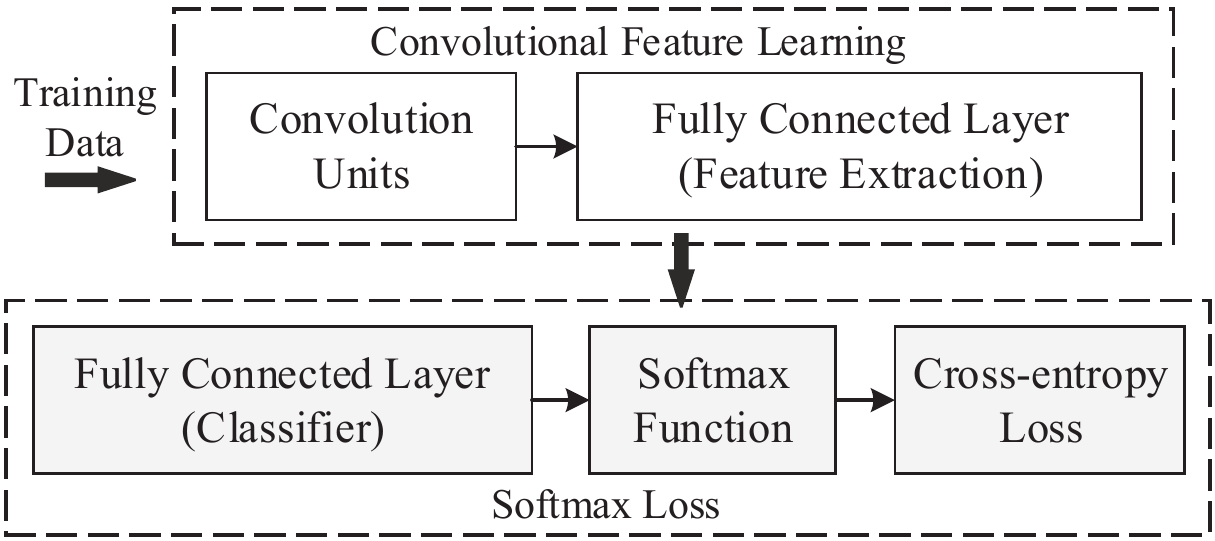}\\
  \caption{Standard CNNs can be viewed as convolutional feature learning machines that are supervised by the softmax loss.}\label{cnn}
  \vspace{-4mm}
\end{figure}

\begin{figure*}
	\centering
	\includegraphics[width=5.6in]{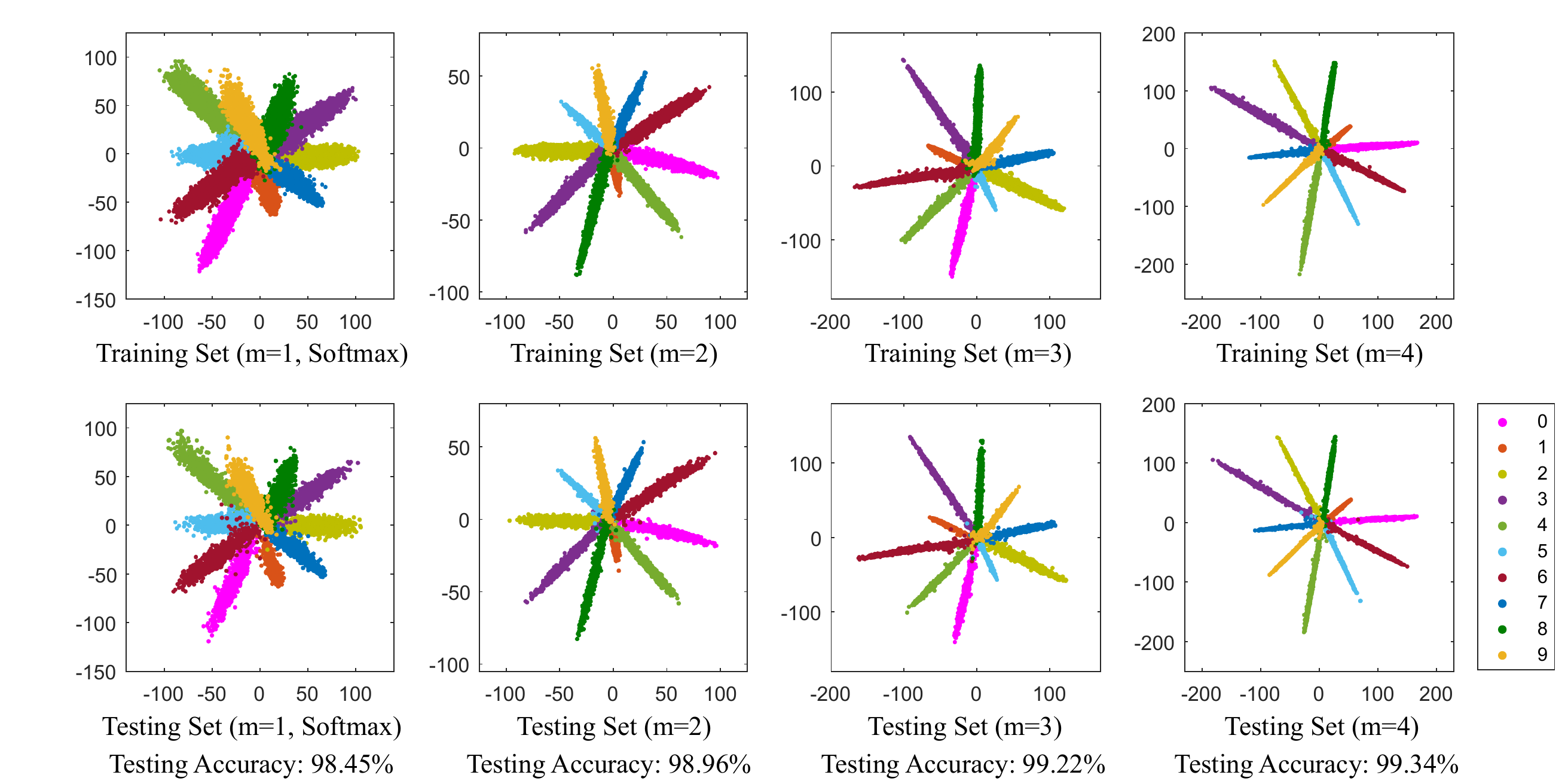}\\
	\vspace{-2mm}
	\caption{CNN-leanrned features visualization (Softmax Loss (m=1) vs. L-Softmax loss (m=2,3,4)) in MNIST dataset. Specifically, we set the feature (input of the L-Softmax loss) dimension as 2, and then plot them by class. We omit the constant term in the fully connected layer, since it just complicates our analysis and nearly does not affect the performance. Note that, the reason why the testing accuracy is not as good as in Table. \ref{mnist} is that we only use 2D features to classify the digits here.}\label{fig1}
	\vspace{-1.5mm}
\end{figure*}

\par
A recent trend towards learning with even stronger features is to reinforce CNNs with more discriminative information. Intuitively, the learned features are good if their intra-class compactness and inter-class separability are simultaneously maximized. While this may not be easy due to the inherent large intra-class variations in many tasks, the strong representation ability of CNNs make it possible to learn invariant features towards this direction. Inspired by such idea, the contrastive loss \cite{hadsell2006dim} and triplet loss \cite{schroff2015facenet} were proposed to enforce extra intra-class compactness and inter-class separability. A consequent problem, however, is that the number of training pairs and triplets can theoretically go up to $\mathcal{O}(N^2)$ where $N$ is the total number of training samples. Considering that CNNs often handle large-scale training sets, a subset of training samples need to be carefully selected for these losses.
The softmax function is widely adopted by many CNNs \cite{krizhevsky2012imagenet,he2015deep,he2015delving} due to its simplicity and probabilistic interpretation. Together with the cross-entropy loss, they form arguably one of the most commonly used components in CNN architectures. In this paper, \textbf{we define the softmax loss as the combination of a cross-entropy loss, a softmax function and the last fully connected layer} (see Fig.~\ref{cnn}). Under such definition, many prevailing CNN models can be viewed as the combination of a convolutional feature learning component and a softmax loss component, as shown in Fig.~\ref{cnn}. Despite its popularity, current softmax loss does not explicitly encourage intra-class compactness and inter-class-separability. Our key intuition is that the separability between sample and parameter can be factorized into amplitude ones and angular ones with cosine similarity: $\bm{W}_c\bm{x}=\|\bm{W}_c\|_2\|\bm{x}\|_2\cos(\theta_c)$, where $c$ is the class index, and the corresponding parameters $\bm{W}_c$ of the last fully connected layer can be regarded as the linear classifier of class $c$. Under softmax loss, the label prediction decision rule is largely determined by the angular similarity to each class since softmax loss uses cosine distance as classification score. The purpose of this paper, therefore, is to generalize the softmax loss to a more general large-margin softmax (L-Softmax) loss in terms of angular similarity, leading to potentially larger angular separability between learned features. This is done by incorporating a preset constant $m$ multiplying with the angle between sample and the classifier of ground truth class. $m$ determines the strength of getting closer to the ground truth class, producing an angular margin. One shall see, the conventional softmax loss becomes a special case of the L-Softmax loss under our proposed framework. Our idea is verified by Fig.~\ref{fig1} where the learned features by L-Softmax become much more compact and well separated.
\par
The L-Softmax loss is a flexible learning objective with adjustable inter-class angular margin constraint. It presents a learning task of adjustable difficulty where the difficulty gradually increases as the required margin becomes larger. The L-Softmax loss has several desirable advantages. First, it encourages angular decision margin between classes, generating more discriminative features. Its geometric interpretation is very clear and intuitive, as elaborated in Section 3.2. Second, it partially avoids overfitting by defining a more difficult learning target, casting a different viewpoint to the overfitting problem. Third, L-Softmax benefits not only classification problems, but also verification problems where ideally learned features should have the minimum inter-class distance being greater than the maximum intra-class distance. In this case, learning well separated features can significantly improve the performance.

\par
Our experiments validate that L-Softmax can effectively boost the performance in both classification and verification tasks. More intuitively, the visualizations of the learned features in Fig.~\ref{fig1} and Fig.~\ref{conmat} show great discriminativeness of the L-Softmax loss. As a straightforward generalization of softmax loss, L-Softmax loss not only inherits all merits from softmax loss but also learns features with large angular margin between different classes. Besides that, the L-Softmax loss is also well motivated with clear geometric interpretation as elaborated in Section 3.3.
\setcounter{footnote}{0}
\renewcommand\thefootnote{\arabic{footnote}}
\section{Related Work and Preliminaries}
Current widely used data loss functions in CNNs include Euclidean loss, (square) hinge loss, information gain loss, contrastive loss, triplet loss, Softmax loss, etc. To enhance the intra-class compactness and inter-class separability, \cite{sun2014deep} trains the CNN with the combination of softmax loss and contrastive loss. The contrastive loss inputs the CNNs with pairs of training samples. If the input pair belongs to the same class, the contrastive loss will require their features are as similar as possible. Otherwise, the contrastive loss will require their distance larger than a margin. \cite{schroff2015facenet} uses the triplet loss to encourage a distance constraint similar to the contrastive loss. Differently, the triplet loss requires 3 (or a multiple of 3) training samples as input at a time. The triplet loss minimizes the distance between an anchor sample and a positive sample (of the same identity), and maximizes the distance between the anchor sample and a negative sample (of different identity). Both triplet loss and contrastive loss require a carefully designed pair selection procedure. Both \cite{sun2014deep} and \cite{schroff2015facenet} suggest that enforcing such a distance constraint that encourages intra-class compactness and inter-class separability can greatly boost the feature discriminativeness, which motivates us to employ a margin constraint in the original softmax loss.
\par
Unlike any previous work, our work cast a novel view on generalizing the original softmax loss. We define the $i$-th input feature $\bm{x}_i$ with the label $y_i$. Then the original softmax loss can be written as
\begin{equation}\label{eq1}
\small
L=\frac{1}{N}\sum_iL_i=\frac{1}{N}\sum_i-\log\bigg(\frac{e^{f_{y_i}}}{\sum_je^{f_j}}\bigg)
\end{equation}
where $f_j$ denotes the $j$-th element ($j\in[1,K]$, K is the number of classes) of the vector of class scores $\bm{f}$, and $N$ is the number of training data. In the softmax loss, $\bm{f}$ is usually the activations of a fully connected layer $\bm{W}$, so $f_{y_i}$ can be written as $f_{y_i}=\bm{W}_{y_i}^T\bm{x}_i$ in which $\bm{W}_{y_i}$ is the ${y_i}$-th column of $\bm{W}$. Note that, we omit the constant $b$ in $f_j,\forall j$ here to simplify analysis, but our L-Softmax loss can still be easily modified to work with $b$. (In fact, the performance is nearly of no difference, so we do not make it complicated here.) Because $f_j$ is the inner product between $\bm{W}_j$ and $\bm{x}_i$, it can be also formulated as $f_j=\|\bm{W}_j\|\|\bm{x}_i\|\cos(\theta_j)$ where $\theta_j$ ($0\leq\theta_j\leq\pi$) is the angle between the vector $\bm{W}_j$ and $\bm{x}_i$. Thus the loss becomes
\begin{equation}\label{eq2}
\small
L_i=-\log\bigg(\frac{e^{\|\bm{W}_{y_i}\|\|\bm{x}_i\|\cos(\theta_{y_i})}}{\sum_je^{\|\bm{W}_j\|\|\bm{x}_i\|\cos(\theta_j)}}\bigg)
\end{equation}
\section{Large-Margin Softmax Loss}
\subsection{Intuition}
We give a simple example to describe our intuition. Consider the binary classification and we have a sample $\bm{x}$ from class 1. The original softmax is to force $\bm{W}_1^T\bm{x}>\bm{W}_2^T\bm{x}$ (i.e. $\|\bm{W}_1\|\|\bm{x}\|\cos(\theta_1)>\|\bm{W}_2\|\|\bm{x}\|\cos(\theta_2)$) in order to classify $\bm{x}$ correctly. However, we want to make the classification more rigorous in order to produce a decision margin. So we instead require $\|\bm{W}_1\|\|\bm{x}\|\cos(m\theta_1)>\|\bm{W}_2\|\|\bm{x}\|\cos(\theta_2)$ ($0\leq\theta_1\leq\frac{\pi}{m}$) where $m$ is a positive integer. Because the following inequality holds:
\begin{equation}\label{ineq}
\small
\begin{aligned}
\|\bm{W}_1\|\|\bm{x}\|\cos(\theta_1)&\geq\|\bm{W}_1\|\|\bm{x}\|\cos(m\theta_1)\\
&>\|\bm{W}_2\|\|\bm{x}\|\cos(\theta_2).
\end{aligned}
\end{equation}
Therefore, $\|\bm{W}_1\|\|\bm{x}\|\cos(\theta_1)>\|\bm{W}_2\|\|\bm{x}\|\cos(\theta_2)$ has to hold. So the new classification criteria is a stronger requirement to correctly classify $\bm{x}$, producing a more rigorous decision boundary for class 1.
\subsection{Definition}
Following the notation in the preliminaries, the L-Softmax loss is defined as
\begin{equation}\label{gsl}
\small
L_i=
-\log\bigg(\frac{e^{\|\bm{W}_{y_i}\|\|\bm{x}_i\|\psi(\theta_{y_i})}}{
e^{\|\bm{W}_{y_i}\|\|\bm{x}_i\|\psi(\theta_{y_i})}+\sum_{j\neq y_i}e^{\|\bm{W}_j\|\|\bm{x}_i\|\cos(\theta_j)}
}\bigg)
\end{equation}
in which we generally require
\begin{equation}
\small
\psi(\theta)=
\left\{
{\begin{array}{*{20}{l}}
{\cos(m\theta),\ \ \ \ 0\leq\theta\leq\dfrac{\pi}{m}}\\
{\mathcal{D}(\theta),\ \ \ \ \dfrac{\pi}{m}<\theta\leq\pi}
\end{array}} \right.
\end{equation}
where $m$ is a integer that is closely related to the classification margin. With larger $m$, the classification margin becomes larger and the learning objective also becomes harder. Meanwhile, $\mathcal{D}(\theta)$ is required to be a monotonically decreasing function and $\mathcal{D}(\frac{\pi}{m})$ should equal $\cos(\frac{\pi}{m})$.
\par

\begin{figure}[h]
  \centering
  \includegraphics[width=2.6in]{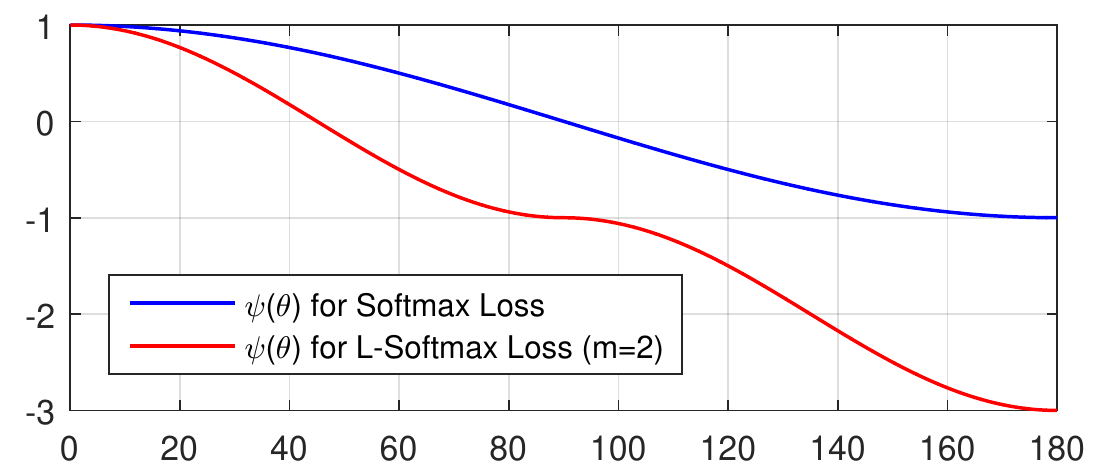}\\
  \vspace{-1.5mm}
  \caption{$\psi(\theta)$ for softmax loss and L-Softmax loss.}\label{fig2}
  \vspace{1mm}
\end{figure}

To simplify the forward and backward propagation, we construct a specific $\psi(\theta_i)$ in this paper:
\begin{equation}\label{gslpsi}
\small
\psi(\theta)=(-1)^k\cos(m\theta)-2k,\ \ \ \theta\in[\frac{k\pi}{m},\frac{(k+1)\pi}{m}]
\end{equation}
where $k\in[0,m-1]$ and $k$ is an integer. Combining Eq. \eqref{eq1}, Eq. \eqref{gsl} and Eq. \eqref{gslpsi}, we have the L-Softmax loss that is used throughout the paper. For forward and backward propagation, we need to replace $\cos(\theta_j)$ with $\frac{\bm{W}_j^T\bm{x}_i}{\|\bm{W}_j\|\|\bm{x}_i\|}$, and replace $\cos(m\theta_{y_i})$ with
\begin{equation}\label{cosmt}
\small
\begin{aligned}
\cos(m\theta_{y_i})&=C_m^0\cos^m(\theta_{y_i})-C_m^2\cos^{m-2}(\theta_{y_i})(1-\cos^{2}(\theta_{y_i}))\\
&+C_m^4\cos^{m-4}(\theta_{y_i})(1-\cos^{2}(\theta_{y_i}))^2+\cdots\\
&(-1)^n C_m^{2n}\cos^{m-2n}(\theta_{y_i})(1-\cos^{2}(\theta_{y_i}))^n+\cdots
\end{aligned}
\end{equation}
where $n$ is an integer and $2n\leq m$. After getting rid of $\theta$, we could perform derivation with respect to $\bm{x}$ and $\bm{W}$. It is also trivial to perform derivation with mini-batch input.
\subsection{Geometric Interpretation}
We aim to encourage aa angle margin between classes via the L-Softmax loss. To simplify the geometric interpretation, we analyze the binary classification case where there are only $\bm{W}_1$ and $\bm{W}_2$.
\par
\begin{figure}[t]
  \centering
  \includegraphics[width=3.3in]{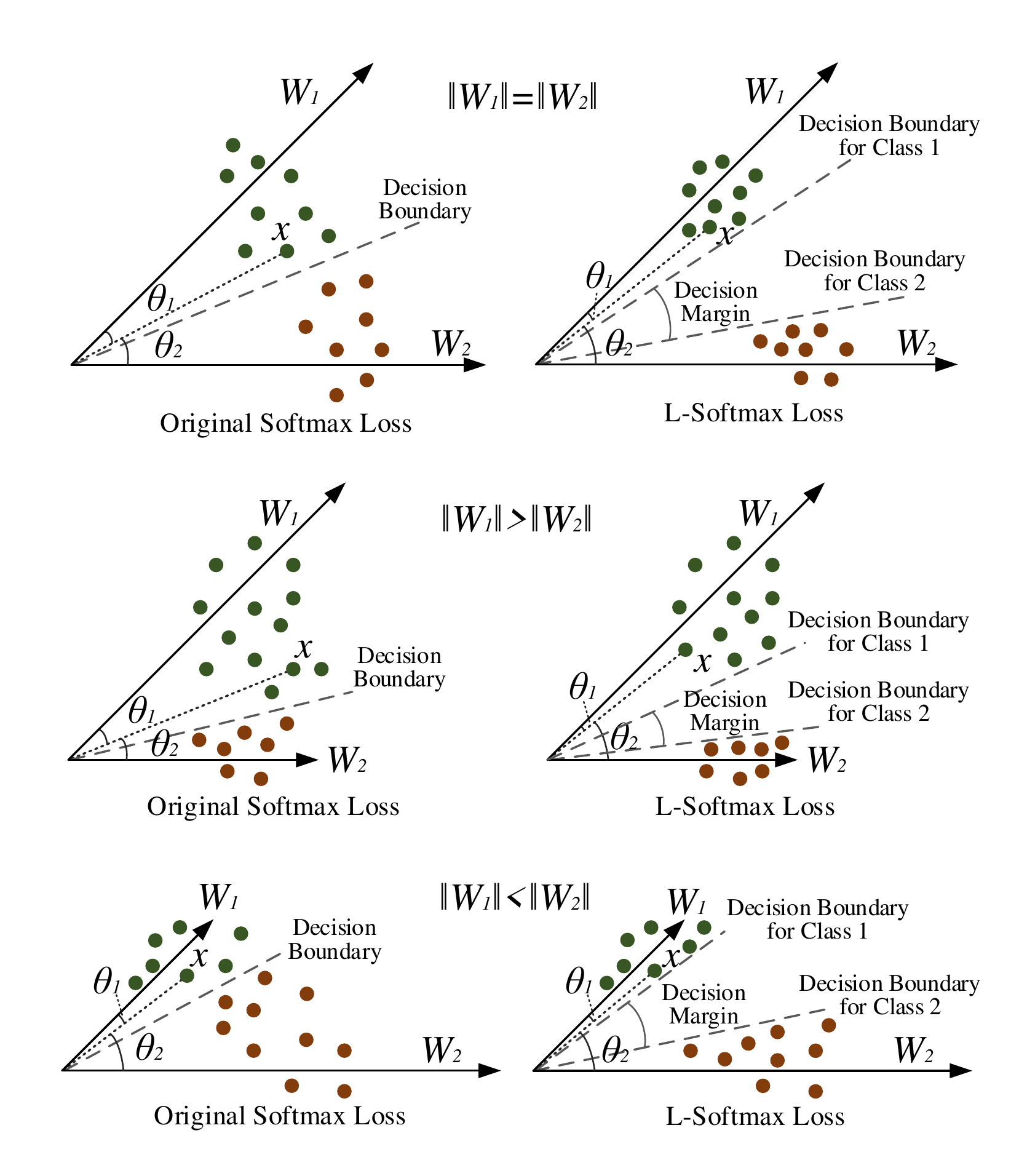}\\
  \caption{Examples of Geometric Interpretation.}\label{fig2}
\end{figure}
First, we consider the $\|\bm{W}_1\|=\|\bm{W}_2\|$ scenario as shown in Fig.~\ref{fig2}. With $\|\bm{W}_1\|=\|\bm{W}_2\|$, the classification result depends entirely on the angles between $\bm{x}$ and $\bm{W}_1$($\bm{W}_2$). In the training stage, the original softmax loss requires $\theta_1<\theta_2$ to classify the sample $\bm{x}$ as class 1, while the L-Softmax loss requires $m\theta_1<\theta_2$ to make the same decision. We can see the L-Softmax loss is more rigor about the classification criteria, which leads to a classification margin between class 1 and class 2. If we assume both softmax loss and L-Softmax loss are optimized to the same value and all training features can be perfectly classified, then the angle margin between class 1 and class 2 is given by $\frac{m-1}{m+1}\theta_{1,2}$ where $\theta_{1,2}$ is the angle between classifier vector $\bm{W}_1$ and $\bm{W}_2$. The L-Softmax loss also makes the decision boundaries for class 1 and class 2 different as shown in Fig \ref{fig2}, while originally the decision boundaries are the same. From another viewpoint, we let $\theta'_1=m\theta_1$ and assume that both the original softmax loss and the L-Softmax loss can be optimized to the same value. Then we can know $\theta'_1$ in the original softmax loss is $m-1$ times larger than $\theta_1$ in the L-Softmax loss. As a result, the angle between the learned feature and $\bm{W}_1$ will become smaller. For every class, the same conclusion holds. In essence, the L-Softmax loss narrows the feasible angle\footnote{Feasible angle of the $i$-th class refers to the possible angle between $\bm{x}$ and $\bm{W}_i$ that is learned by CNNs.} for every class and produces an angle margin between these classes.
\par
For both the $\|\bm{W}_1\|>\|\bm{W}_2\|$ and $\|\bm{W}_1\|<\|\bm{W}_2\|$ scenarios, the geometric interpretation is a bit more complicated. Because the length of $\bm{W}_1$ and $\bm{W}_2$ is different, the feasible angles of class 1 and class 2 are also different (see the decision boundary of original softmax loss in Fig.~\ref{fig2}). Normally, the larger $\bm{W}_j$ is, the larger the feasible angle of its corresponding class is. As a result, the L-Softmax loss also produces different feasible angles for different classes. Similar to the analysis of the $\|\bm{W}_1\|=\|\bm{W}_2\|$ scenario, the proposed loss will also generate a decision margin between class 1 and class 2.
\subsection{Discussion}
The L-Softmax loss utilizes a simple modification over the original softmax loss, achieving a classification angle margin between classes. By assigning different values for $m$, we define a flexible learning task with adjustable difficulty for CNNs. The L-Softmax loss is endowed with some nice properties such as
\begin{itemize}
\item The L-Softmax loss has a clear geometric interpretation. $m$ controls the margin among classes. With bigger $m$ (under the same training loss), the ideal margin between classes becomes larger and the learning difficulty is also increased. With $m=1$, the L-Softmax loss becomes identical to the original softmax loss.
\item The L-Softmax loss defines a relatively difficult learning objective with adjustable margin (difficulty). A difficult learning objective can effectively avoid over-fitting and take full advantage of the strong learning ability from deep and wide architectures.
\item The L-Softmax loss can be easily used as a drop-in replacement for standard loss, as well as used in tandem with other performance-boosting approaches and modules, including learning activation functions, data augmentation, pooling functions or other modified network architectures.
\end{itemize}

\begin{table*}[t]
\centering
\footnotesize
\begin{tabular}{|c|c|c|c|c|c|}
\hline
Layer & MNIST (for Fig.~\ref{fig1}) & MNIST & CIFAR10/CIFAR10+ & CIFAR100 & LFW\\
\hline\hline
Conv0.x & N/A & [3$\times$3, 64]$\times$1 & [3$\times$3, 64]$\times$1  & [3$\times$3, 96]$\times$1 & [3$\times$3, 64]$\times$1, Stride 2 \\\hline
Conv1.x  & [5$\times$5, 32]$\times$2, Padding 2 & [3$\times$3, 64]$\times$3 & [3$\times$3, 64]$\times$4 & [3$\times$3, 96]$\times$4 & [3$\times$3, 64]$\times$4 \\\hline
Pool1&\multicolumn{5}{c|}{2$\times$2 Max, Stride 2}\\\hline
Conv2.x  & [5$\times$5, 64]$\times$2, Padding 2 & [3$\times$3, 64]$\times$3 & [3$\times$3, 96]$\times$4 & [3$\times$3, 192]$\times$4 & [3$\times$3, 256]$\times$4 \\\hline
Pool2 & \multicolumn{5}{c|}{2$\times$2 Max, Stride 2}\\\hline
Conv3.x & [5$\times$5, 128]$\times$2, Padding 2  & [3$\times$3, 64]$\times$3 & [3$\times$3, 128]$\times$4 & [3$\times$3, 384]$\times$4 & [3$\times$3, 256]$\times$4 \\\hline
Pool3 & \multicolumn{5}{c|}{2$\times$2 Max, Stride 2}\\\hline
Conv4.x & N/A & N/A & N/A & N/A & [3$\times$3, 256]$\times$4 \\\hline
Fully Connected & 2 & 256 & 256 & 512 & 512 \\\hline%Loss  & Softmax & Softmax & Softmax & Softmax \\\hline
\end{tabular}
\caption{Our CNN architectures for different benchmark datasets. Conv1.x, Conv2.x and Conv3.x denote convolution units that may contain multiple convolution layers. E.g., [3$\times$3, 64]$\times$4 denotes 4 cascaded convolution layers with 64 filters of size 3$\times$3. }\label{netarch}
\end{table*}

\section{Optimization}
It is easy to compute the forward and backward propagation for the L-Softmax loss, so it is also trivial to optimize the L-Softmax loss using typical stochastic gradient descent. For $L_i$, the only difference between the original softmax loss and the L-Softmax loss lies in $f_{y_i}$. Thus we only need to compute $f_{y_i}$ in forward and backward propagation while $f_j,j\neq {y_i}$ is the same as the original softmax loss. Putting in Eq. \eqref{gslpsi} and Eq. \eqref{cosmt}, $f_{y_i}$ is written as
\begin{equation}\label{fi}
\small
\begin{aligned}
f_{y_i}=&(-1)^k\cdot\|\bm{W}_{y_i}\|\|\bm{x}_i\| \cos(m\theta_i)-2k\cdot\|\bm{W}_{y_i}\|\|\bm{x}_i\|\\
=&(-1)^k\cdot\|\bm{W}_{y_i}\|\|\bm{x}_i\| \bigg( C_m^0\big(\frac{\bm{W}_{y_i}^T\bm{x}_i}{\|\bm{W}_{y_i}\|\|\bm{x}_i\|}\big)^m-\\ &C_m^2\big(\frac{\bm{W}_{y_i}^T\bm{x}_i}{\|\bm{W}_{y_i}\|\|\bm{x}_i\|}\big)^{m-2}(1-\big(\frac{\bm{W}_{y_i}^T\bm{x}_i}{\|\bm{W}_{y_i}\|\|\bm{x}_i\|}\big)^{2})+\cdots\bigg)\\
&-2k\cdot\|\bm{W}_{y_i}\|\|\bm{x}_i\|
\end{aligned}
\end{equation}
where $\frac{\bm{W}_{y_i}^T\bm{x}}{\|\bm{W}_{y_i}\|\|\bm{x}\|}\in[\cos(\frac{k\pi}{m}),\cos(\frac{(k+1)\pi}{m})]$ and $k$ is an integer that belongs to $[0,m-1]$. For the backward propagation, we use the chain rule to compute the partial derivative: $\frac{\partial L_i}{\partial \bm{x}_i}=\sum_j\frac{\partial L_i}{\partial f_{j}}\frac{\partial f_{j}}{\partial \bm{x}_i}$ and $\frac{\partial L_i}{\partial \bm{W}_{y_i}}=\sum_j\frac{\partial L_i}{\partial f_{j}}\frac{\partial f_{j}}{\partial \bm{W}_{y_i}}$. Because $\frac{\partial L_i}{\partial f_{j}}$ and $\frac{\partial f_j}{\partial \bm{x}_{i}},\frac{\partial f_j}{\partial \bm{W}_{y_i}}, \forall j \neq y_i$ are the same for both original softmax loss and L-Softmax loss, we leave it out for simplicity. $\frac{\partial f_{y_i}}{\partial \bm{x}_i}$ and $\frac{\partial f_{y_i}}{\partial \bm{W}_{y_i}}$ can be computed via
\begin{equation}\label{pardevx}
\small
\begin{aligned}
&\frac{\partial f_{y_i}}{\partial \bm{x}_i}=(-1)^k\cdot\bigg( C_m^0\frac{m(\bm{W}_{y_i}^T\bm{x}_i)^{m-1}\bm{W}_{y_i}}{(\|\bm{W}_{y_i}\|\|\bm{x}_i\|)^{m-1}}-\\
&\ \ \ C_m^0\frac{(m-1)(\bm{W}_{y_i}^T\bm{x}_i)^m\bm{x}_i}{\|\bm{W}_{y_i}\|^{m-1}\|\bm{x}_i\|^{m+1}}-C_m^2\frac{(m-2)(\bm{W}_{y_i}^T\bm{x}_i)^{m-3}\bm{W}_{y_i}}{(\|\bm{W}_{y_i}\|\|\bm{x}_i\|)^{m-3}}\\ &\ \ \ +C_m^2\frac{(m-3)(\bm{W}_{y_i}^T\bm{x}_i)^{m-2}\bm{x}_i}{\|\bm{W}_{y_i}\|^{m-3}\|\bm{x}_i\|^{m-1}}+C_m^2\frac{m(\bm{W}_{y_i}^T\bm{x}_i)^{m-1}\bm{W}_{y_i}}{(\|\bm{W}_{y_i}\|\|\bm{x}_i\|)^{m-1}}\\
&\ \ \ -C_m^2\frac{(m-1)(\bm{W}_{y_i}^T\bm{x}_i)^m\bm{x}_i}{\|\bm{W}_{y_i}\|^{m-1}\|\bm{x}_i\|^{m+1}}+\cdots\bigg)-2k\cdot\frac{\|\bm{W}_{y_i}\|\bm{x}_i}{\|\bm{x}_i\|},\\
\end{aligned}
\end{equation}
\begin{equation}\label{pardevw}
\small
\begin{aligned}
&\frac{\partial f_{y_i}}{\partial \bm{W}_{y_i}}=(-1)^k\cdot\bigg( C_m^0\frac{m(\bm{W}_{y_i}^T\bm{x}_i)^{m-1}\bm{x}_i}{(\|\bm{W}_{y_i}\|\|\bm{x}_i\|)^{m-1}}-\\
&\ \ \ C_m^0\frac{(m-1)(\bm{W}_{y_i}^T\bm{x}_i)^m\bm{W}_{y_i}}{\|\bm{W}_{y_i}\|^{m+1}\|\bm{x}_i\|^{m-1}}-C_m^2\frac{(m-2)(\bm{W}_{y_i}^T\bm{x}_i)^{m-3}\bm{x}_i}{(\|\bm{W}_{y_i}\|\|\bm{x}_i\|)^{m-3}}\\ &\ \ \ +C_m^2\frac{(m-3)(\bm{W}_{y_i}^T\bm{x}_i)^{m-2}\bm{W}_{y_i}}{\|\bm{W}_{y_i}\|^{m-1}\|\bm{x}_i\|^{m-3}}+C_m^2\frac{m(\bm{W}_{y_i}^T\bm{x}_i)^{m-1}\bm{x}_i}{(\|\bm{W}_{y_i}\|\|\bm{x}_i\|)^{m-1}}\\
&\ \ \ -C_m^2\frac{(m-1)(\bm{W}_{y_i}^T\bm{x}_i)^m\bm{W}_{y_i}}{\|\bm{W}_{y_i}\|^{m+1}\|\bm{x}_i\|^{m-1}}+\cdots\bigg)-2k\cdot\frac{\|\bm{x}_i\|\bm{W}_{y_i}}{\|\bm{W}_{y_i}\|}.\\
\end{aligned}
\end{equation}
\par
In implementation, $k$ can be efficiently computed by constructing a look-up table for $\frac{\bm{W}_{y_i}^T\bm{x}_i}{\|\bm{W}_{y_i}\|\|\bm{x}_i\|}$ (i.e. $\cos(\theta_{y_i})$). To be specific, we give an example of the forward and backward propagation when $m=2$. Thus $f_i$ is written as
\begin{equation}\label{fi2}
\small
f_i=(-1)^k \frac{2(\bm{W}_{y_i}^T\bm{x}_i)^2}{\|\bm{W}_{y_i}\|\|\bm{x}_i\|}-\big(2k+(-1)^k\big)\|\bm{W}_{y_i}\|\|\bm{x}_i\|
\end{equation}
where,
$
\small
k=
\left\{
{\begin{array}{*{20}{l}}
{1,\ \ \ \ \frac{\bm{W}_{y_i}^T\bm{x}_i}{\|\bm{W}_{y_i}\|\|\bm{x}_i\|}\leq\cos(\frac{\pi}{2})}\\
{0,\ \ \ \ \frac{\bm{W}_{y_i}^T\bm{x}_i}{\|\bm{W}_{y_i}\|\|\bm{x}_i\|}>\cos(\frac{\pi}{2})}
\end{array}} \right..
$
\par
In the backward propagation, $\frac{\partial f_{y_i}}{\partial \bm{x}_i}$ and $\frac{\partial f_{y_i}}{\partial \bm{W}_{y_i}}$ can be computed with
\begin{equation}\label{devfi2x}
\small
\begin{aligned}
\frac{\partial f_{y_i}}{\partial \bm{x}_i}=&(-1)^k \bigg( \frac{4\bm{W}_{y_i}^T\bm{x}_i\bm{W}_{y_i}}{\|\bm{W}_{y_i}\|\|\bm{x}_i\|}- \frac{2(\bm{W}_{y_i}^T\bm{x}_i)^2\bm{x}_i}{\|\bm{W}_{y_i}\|\|\bm{x}_i\|^3} \bigg)\\
&-\big(2k+(-1)^k\big)\frac{\|\bm{W}_{y_i}\|\bm{x}_i}{\|\bm{x}_i\|},
\end{aligned}
\end{equation}
\begin{equation}\label{devfi2w}
\small
\begin{aligned}
\frac{\partial f_{y_i}}{\partial \bm{W}_{y_i}}=&(-1)^k \bigg( \frac{4\bm{W}_{y_i}^T\bm{x}_i\bm{x}_i}{\|\bm{W}_{y_i}\|\|\bm{x}_i\|}- \frac{2(\bm{W}_{y_i}^T\bm{x}_i)^2\bm{W}_{y_i}}{\|\bm{x}_i\|\|\bm{W}_{y_i}\|^3} \bigg)\\
&-\big(2k+(-1)^k\big)\frac{\|\bm{x}_i\|\bm{W}_{y_i}}{\|\bm{W}_{y_i}\|}.
\end{aligned}
\end{equation}

While $m\geq3$, we can still use Eq. \eqref{fi}, Eq. \eqref{pardevx} and Eq. \eqref{pardevw} to compute the forward and backward propagation.
\par
\begin{figure*}
  \centering
  \includegraphics[width=5.1in]{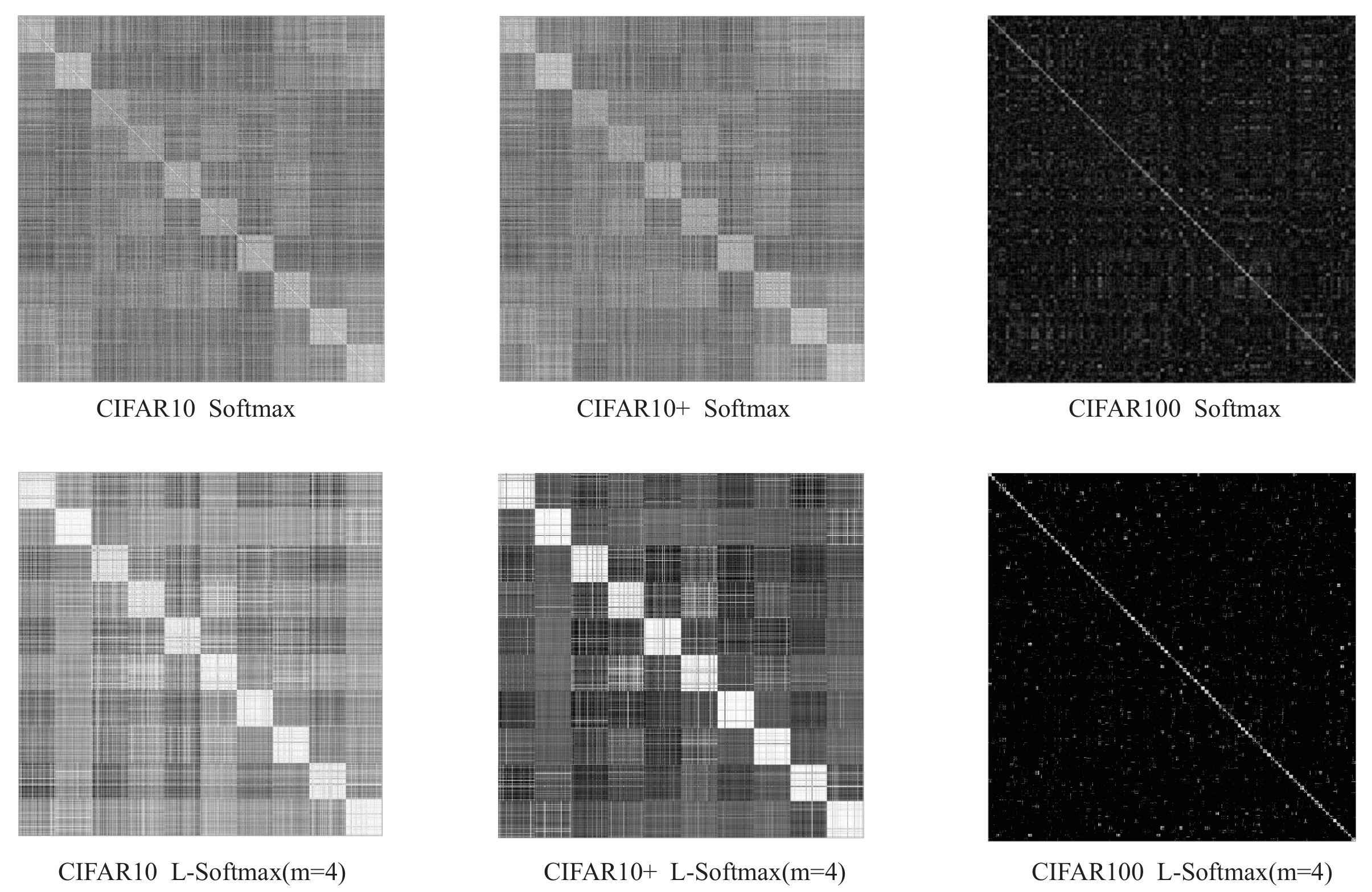}\\
  \caption{Confusion matrix on CIFAR10, CIFAR10+ and CIFAR100.}\label{conmat}
\end{figure*}
\section{Experiments and Results}
\subsection{Experimental Settings}
We evaluate the generalized softmax loss in two typical vision applications: visual classification and face verification. In visual classification, we use three standard benchmark datasets: MNIST \cite{lecun1998mnist}, CIFAR10 \cite{krizhevsky2009learning}, and CIFAR100 \cite{krizhevsky2009learning}. In face verification, we evaluate our method on the widely used LFW dataset \cite{huang2007labeled}. We only use a single model in all baseline CNNs to compare our performance. For convenience, we use L-Softmax to denote the L-Softmax loss. Both Softmax and L-Softmax in the experiments use the same CNN shown in Table \ref{netarch}.
\par
\textbf{General Settings}: We follow the design philosophy of VGG-net \cite{simonyan2014very} in two aspects: (1) for convolution layers, the kernel size is 3$\times$3 and 1 padding (if not specified) to keep the feature map unchanged. (2) for pooling layers, if the feature map size is halved, the number of filters is doubled in order to preserve the time complexity per layer. Our CNN architectures are described in Table~\ref{netarch}. In convolution layers, the stride is set to 1 if not specified. We implement the CNNs using the Caffe library \cite{jia2014caffe} with our modifications. For all experiments, we adopt the PReLU \cite{he2015delving} as the activation functions, and the batch size is 256. We use a weight decay of 0.0005 and momentum of 0.9. The weight initialization in \cite{he2015delving} and batch normalization \cite{ioffe2015batch} are used in our networks but without dropout. Note that we only perform the mean substraction preprocessing for training and testing data. For optimization, normally the stochastic gradient descent will work well. However, when training data has too many subjects (such as CASIA-WebFace dataset), the convergence of L-Softmax will be more difficult than softmax loss. For those cases that L-Softmax has difficulty converging, we use a learning strategy by letting $f_{y_i}=\frac{\lambda\|\bm{W}_{y_i}\|\|\bm{x}_i\|\cos(\theta_{y_i})+\|\bm{W}_{y_i}\|\|\bm{x}_i\|\psi({\theta_{y_i}})}{1+\lambda}$ and start the gradient descent with a very large $\lambda$ (it is similar to optimize the original softmax). Then we gradually reduce $\lambda$ during iteration. Ideally $\lambda$ can be gradually reduced to zero, but in practice, a small value will usually suffice.
\par
\textbf{MNIST, CIFAR10, CIFAR100}: We start with a learning rate of 0.1, divide it by 10 at 12k and 15k iterations, and eventually terminate training at 18k iterations, which is determined on a 45k/5k train/val split.
\par
\textbf{Face Verification}: The learning rate is set to 0.1, 0.01, 0.001 and is switched when the training loss plateaus. The total number of epochs is about is about 30 for our models.
\par
\textbf{Testing}: we use the softmax to classify the testing samples in MNIST, CIFAR10 and CIFAR100 dataset. In LFW dataset, we use the simple cosine distance and the nearest neighbor rule for face verification.
\subsection{Visual Classification}
\textbf{MNIST}: Our network architecture is shown in Table \ref{netarch}. Table \ref{mnist} shows the previous best results and those for our proposed L-Softmax loss. From the results, the L-Softmax loss not only outperforms the original softmax loss using the same network but also achieves the state-of-the-art performance compared to the other deep CNN architectures. In Fig.~\ref{fig1}, we also visualize the learned features by the L-Softmax loss and compare them to the original softmax loss. Fig.~\ref{fig1} validates the effectiveness of the large margin constraint within L-Softmax loss. With larger $m$, we indeed obtain a larger angular decision margin.
\begin{table}[t]
\centering
\footnotesize
\begin{tabular}{|c|c|}
\hline
Method & Error Rate \\
\hline\hline
\qihao{CNN \cite{jarrett2009best}} & 0.53 \\
\qihao{DropConnect \cite{wan2013regularization}} & 0.57\\
\qihao{FitNet \cite{romero2015fitnets}}  & 0.51 \\
\qihao{NiN \cite{lin2013netowrk}}  & 0.47\\
\qihao{Maxout \cite{goodfellow2013maxout}}  & 0.45 \\
\qihao{DSN \cite{lee2015deeply}}  & 0.39 \\
\qihao{R-CNN \cite{liang2015recurrent}} & \textbf{0.31}\\
\qihao{GenPool \cite{lee2015generalizing}} & \textbf{0.31}\\\hline\hline
\qihao{Hinge Loss} & 0.47\\
\qihao{Softmax} & 0.40\\
\qihao{L-Softmax (m=2)} & 0.32\\
\qihao{L-Softmax (m=3)} & \textbf{0.31}\\
\qihao{L-Softmax (m=4)} & \textbf{0.31}\\
\hline
\end{tabular}
\caption{Recognition error rate (\%) on MNIST dataset.}\label{mnist}
\end{table}
\par
\textbf{CIFAR10}: We use two commonly used comparison protocols in CIFAR10 dataset. We first compare our L-Softmax loss under no data augmentation setup. For the data augmentation experiment, we follow the standard data augmentation in \cite{lee2015deeply} for training: 4 pixels are padded on each side, and a 32$\times$32 crop is randomly sampled from the padded image or its horizontal flip. In testing, we only evaluate the single view of the original 32$\times$32 image. The results are shown in Table \ref{cifar10}. One can observe that our L-Softmax loss greatly boosts the accuracy, achieving 1\%-2\% improvement over the original softmax loss and the other state-of-the-art CNNs.

\begin{table}[t]
\centering
\footnotesize
\begin{tabular}{|c|c|c|}
\hline
Method & CIFAR10 & CIFAR10+ \\
\hline\hline
\qihao{DropConnect \cite{wan2013regularization}} & 9.41 & 9.32\\
\qihao{FitNet \cite{romero2015fitnets}}  & N/A & 8.39\\
\qihao{NiN + LA units \cite{lin2013netowrk}}  & 10.47 & 8.81\\
\qihao{Maxout \cite{goodfellow2013maxout}}  & 11.68 & 9.38\\
\qihao{DSN \cite{lee2015deeply}}  & 9.69 & 7.97\\
\qihao{All-CNN \cite{springenberg2015striving}} & 9.08 & 7.25\\
\qihao{R-CNN \cite{liang2015recurrent}} & 8.69 & 7.09\\
\qihao{ResNet \cite{he2015deep}} & N/A & 6.43\\
\qihao{GenPool \cite{lee2015generalizing}} &7.62 & 6.05\\\hline\hline
\qihao{Hinge Loss} & 9.91 & 6.96\\
\qihao{Softmax} & 9.05 & 6.50\\
\qihao{L-Softmax (m=2)} & 7.73 & 6.01\\
\qihao{L-Softmax (m=3)} & 7.66 & 5.94\\
\qihao{L-Softmax (m=4)} & \textbf{7.58} & \textbf{5.92}\\
\hline
\end{tabular}
\caption{Recognition error rate (\%) on CIFAR10 dataset. CIFAR10 denotes the performance without data augmentation, while CIFAR10+ is with data augmentation.}\label{cifar10}
\end{table}
\par
\textbf{CIFAR100}: We also evaluate the generalize softmax loss on the CIFAR100 dataset. The CNN architecture refers to Table \ref{netarch}. One can notice that the L-Softmax loss outperform the CNN with softmax loss and all the other competitive methods. The L-Softmax loss improves more than 2.5\% accuracy over the CNN and more than 1\% over the current state-of-the-art CNN.
\par
\textbf{Confusion Matrix Visualization}: We also give the confusion matrix comparison between the softmax baseline and the L-Softmax loss (m=4) in Fig.~\ref{conmat}. Specifically we normalize the learned features and then calculate the cosine distance between these features. From Fig.~\ref{conmat}, one can see that the intra-class compactness is greatly enhanced while the inter-class separability is also enlarged.
\par
\begin{table}[t]
\centering
\footnotesize
\begin{tabular}{|c|c|}
\hline
Method & Error Rate \\
\hline\hline
\qihao{FitNet \cite{romero2015fitnets}}  & 35.04 \\
\qihao{NiN \cite{lin2013netowrk}}  & 35.68\\
\qihao{Maxout \cite{goodfellow2013maxout}}  & 38.57 \\
\qihao{DSN \cite{lee2015deeply}}  & 34.57 \\
\qihao{dasNet \cite{stollenga2014deep}} & 33.78\\
\qihao{All-CNN \cite{springenberg2015striving}}  & 33.71 \\
\qihao{R-CNN \cite{liang2015recurrent}} & 31.75\\
\qihao{GenPool \cite{lee2015generalizing}} & 32.37\\\hline\hline
\qihao{Hinge Loss} & 32.90\\
\qihao{Softmax} & 32.74\\
\qihao{L-Softmax (m=2)} & 29.95\\
\qihao{L-Softmax (m=3)} & 29.87\\
\qihao{L-Softmax (m=4)} & \textbf{29.53}\\
\hline
\end{tabular}
\caption{Recognition error rate (\%) on CIFAR100 dataset.}\label{cifar100}
\end{table}
\begin{table}[t]
\centering
\footnotesize
\begin{tabular}{|c|c|c|}
\hline
Method & Outside Data & Accuracy \\
\hline\hline
\qihao{FaceNet \cite{schroff2015facenet}} & 200M* & \textbf{99.65} \\
\qihao{Deep FR \cite{parkhi2015deep}} & 2.6M & 98.95\\
\qihao{DeepID2+ \cite{sun2015deeply}} & 300K* & 98.70 \\\hline\hline
\qihao{\cite{yi2014learning}} & WebFace & 97.73 \\
\qihao{\cite{ding2015robust}} & WebFace & 98.43 \\\hline\hline
\qihao{Softmax} & WebFace & 96.53\\
\qihao{Softmax + Contrastive} & WebFace & 97.31\\
\qihao{L-Softmax (m=2)} & WebFace & 97.81\\
\qihao{L-Softmax (m=3)} & WebFace & 98.27\\
\qihao{L-Softmax (m=4)} & WebFace & \textbf{98.71}\\
\hline
\end{tabular}
\caption{Verification performance (\%) on LFW dataset. * denotes the outside data is private (not publicly available).}\label{lfw}
\end{table}

\begin{figure*}[t]
	\centering
	\includegraphics[width=5.4in]{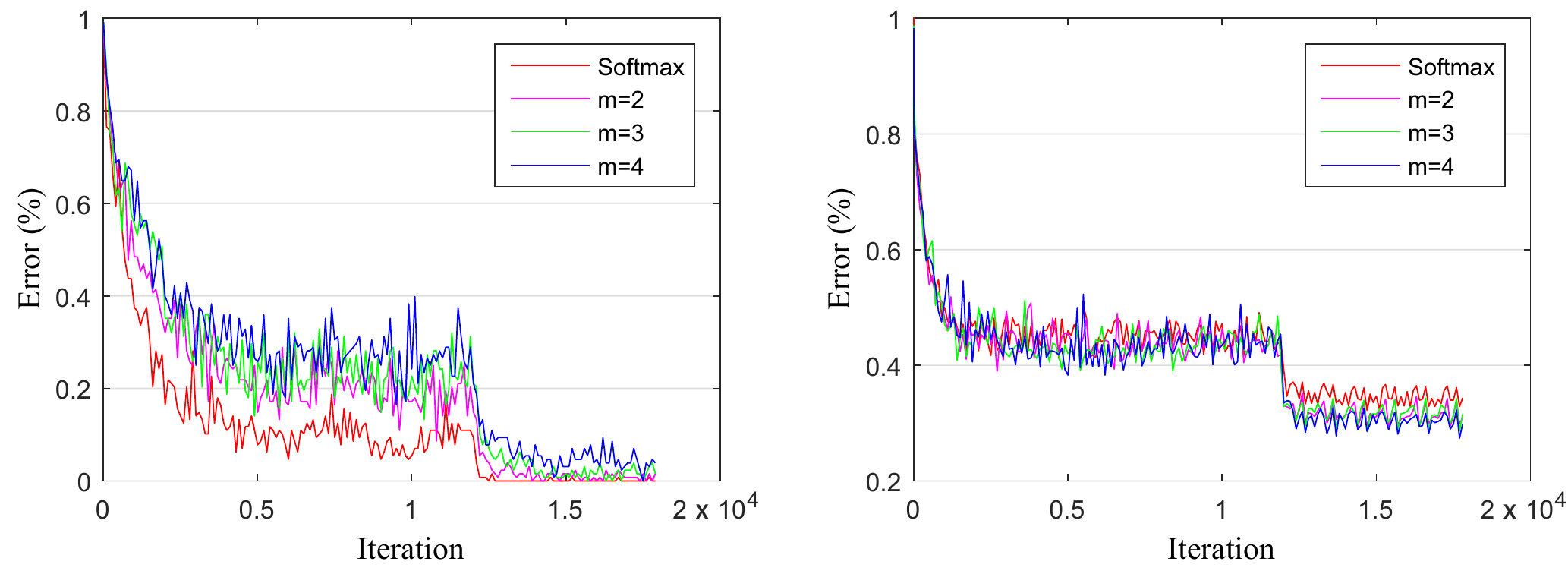}\\
	\vspace{-1.5mm}
	\caption{Error vs. iteration with different value of $m$ on CIFAR100. The left shows training error and the right shows testing error.}\label{diffm}
	\vspace{-1mm}
\end{figure*}
\begin{figure*}[t]
	\centering
	\includegraphics[width=5.4in]{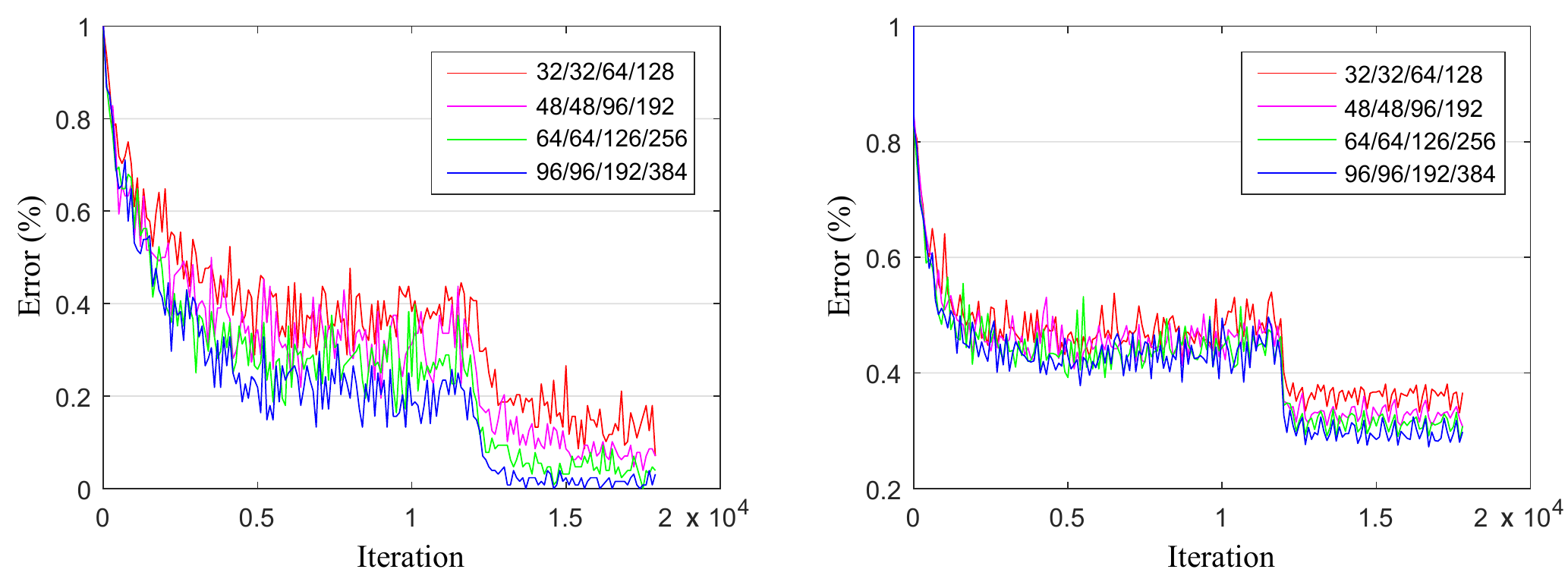}\\
	\vspace{-1.5mm}
	\caption{Error vs. iteration (m=4) with different number of filters on CIFAR100. The left (right) presents training (testing) error.}\label{diffk}
	\vspace{-1mm}
\end{figure*}

\textbf{Error Rate vs. Iteration}: Fig.~\ref{diffm} illustrates the relation between the error rate and the iteration number with different $m$ in the L-Softmax loss. We use the same CNN (same as the CIFAR10 network) to optimize the L-Softmax loss with $m=1,2,3,4$, and then plot their training and testing error rate. One can observe that the original softmax suffers from severe overfitting problem (training loss is very low but testing loss is higher), while the L-Softmax loss can greatly avoid such problem. Fig.~\ref{diffk} shows the relation between the error rate and the iteration number with different number of filters in the L-Softmax loss (m=4). We use four different CNN architecture to optimize the L-Softmax loss with $m=4$, and then plot their training and testing error rate. These four CNN architectures have the same structure and only differ in the number of filters (e.g. 32/32/64/128 denotes that there are 32, 32, 64 and 128 filters in every convolution layer of Conv0.x, Conv1.x Conv2.x and Conv3.x, respectively). On both the training set and testing set, the L-Softmax loss with larger number of filters performs better than those with smaller number of filters, indicating L-Softmax loss does not easily suffer from overfitting. The results also show that our L-Softmax loss can be optimized easily. Therefore, one can learn that the L-Softmax loss can make full use of the stronger learning ability of CNNs, since stronger learning ability leads to performance gain.

\subsection{Face Verification}
To further evaluate the learned features, we conduct an experiment on the famous LFW dataset \cite{huang2007labeled}. The dataset collects 13,233 face images from 5749 persons from uncontrolled conditions. Following the unrestricted with labeled outside data protocol \cite{huang2007labeled}, we train on the publicly available CASIA-WebFace \cite{yi2014learning} outside dataset (490k labeled face images belonging to over 10,000 individuals) and test on the 6,000 face pairs on LFW. People overlapping between the outside training data and the LFW testing data are excluded. As preprocessing, we use IntraFace \cite{asthana2014incremental} to align the face images and then crop them based on 5 points. Then we train a single network for feature extraction, so we only compare the single model performance of current state-of-the-art CNNs. Finally PCA is used to form a compact feature vector. The results are given in Table \ref{lfw}. The generalize softmax loss achieves the current best results while only trained with the CASIA-WebFace outside data, and is also comparable to the current state-of-the-art CNNs with private outside data. Experimental results well validate the conclusion that the L-Softmax loss encourages the intra-class compactness and inter-class separability.
\vspace{-1.7mm}
\section{Concluding Remarks}
We proposed the Large-Margin Softmax loss for the convolutional neural networks. The large-margin softmax loss defines a flexible learning task with adjustable margin. We can set the parameter $m$ to control the margin. With larger $m$, the decision margin between classes also becomes larger. More appealingly, the Large-Margin Softmax loss has very clear intuition and geometric interpretation. The extensive experimental results on several benchmark datasets show clear advantages over current state-of-the-art CNNs and all the compared baselines.

\section*{Acknowledgement}
The authors would like to thank Prof. Le Song (Georgia Tech) for constructive suggestions. This work is partially supported by the National Natural Science Foundation for Young Scientists of China (Grant no.61402289) and National Science Foundation of Guangdong Province (Grant no. 2014A030313558).
\nocite{langley00}
\par
{
\small
\bibliography{example_paper}
\bibliographystyle{icml2016}
}
\end{document}